\PassOptionsToPackage{table}{xcolor}

\documentclass[runningheads]{llncs}

\usepackage{eccv}

\usepackage{eccvabbrv}      % ECCV standard abbreviations (\eg, \etal, etc.)
\usepackage{graphicx}       % Figure support
\usepackage{booktabs}       % Professional table rules
\usepackage{multirow}       % Multi-row tables
\usepackage{adjustbox}      % Adjust tables/figures
\usepackage[accsupp]{axessibility} % Improve PDF accessibility
\usepackage{makecell}

\usepackage{amsfonts}
\usepackage{amsmath}
\usepackage{nicefrac}
\usepackage{pifont}

\usepackage{verbatim}
\usepackage{wrapfig}
\usepackage{marvosym}
\usepackage{enumitem}

\definecolor{cornellred}{rgb}{0.7, 0.11, 0.11}
\definecolor{cadmiumgreen}{rgb}{0.0, 0.42, 0.24}
\definecolor{aliceblue}{rgb}{0.91, 0.94, 0.97}
\definecolor{darkblue}{rgb}{0.83, 0.89, 0.97}
\definecolor{Red7}{rgb}{0.941, 0.243, 0.243}
\definecolor{Green7}{RGB}{55, 178, 77}
\definecolor{Indigo}{RGB}{75, 0, 130}
\definecolor{Blue9}{rgb}{0.098,0.3,0.9}
\definecolor{think}{HTML}{B5739D}
\definecolor{omni}{HTML}{67AB9F}
\definecolor{Blue1}{HTML}{0000CC}
\definecolor{deepgreen}{HTML}{008000}
\definecolor{lightgreen}{rgb}{0.9, 1, 0.9}

\usepackage{hyperref}

\usepackage{orcidlink}

\begin{document}

% ---------------------------------------------------------------
% Title
% ---------------------------------------------------------------
\title{Video Streaming Thinking: VideoLLMs Can Watch and Think Simultaneously}

\titlerunning{Video Streaming Thinking}

% ---------------------------------------------------------------
% Authors
% Include ORCID if possible.
% Replace 0000-0000-0000-0000 with real ORCID IDs.
% ---------------------------------------------------------------
\author{
Yiran Guan\inst{1*}\orcidlink{0009-0000-3407-7581} \and
Liang Yin\inst{1*}\orcidlink{0009-0005-4218-0323} \and
Dingkang Liang\inst{1}\orcidlink{0000-0003-3035-1373} \and
Jianzhong Ju\inst{2}\orcidlink{0009-0001-3453-9800} \and \\
Zhenbo Luo\inst{2}\orcidlink{0009-0002-5836-0749} \and
Jian Luan\inst{2}\orcidlink{0000-0002-2383-226X} \and
Yuliang Liu\inst{1}\orcidlink{0000-0002-3037-173X} \and
Xiang Bai\inst{1\dag}\orcidlink{0000-0002-3449-5940}
}

% If there are more than two authors, "et al." is used.
\authorrunning{Y.~Guan et al.}

% ---------------------------------------------------------------
% Institutes
% ---------------------------------------------------------------
\institute{
Huazhong University of Science and Technology, Wuhan, China\\
\and
MiLM Plus, Xiaomi Inc., Beijing, China
\email{\{yiranguan,liangyin,dkliang,xbai\}@hust.edu.cn}
}

\maketitle

% Optional author note.
% Use this only if ECCV/Springer allows footnotes for equal contribution/corresponding author.
% This creates an unnumbered footnote after the title block.
{\let\thefootnote\relax
\footnotetext{
*~Equal contribution. 
\dag~Corresponding author.
}
}

\begin{abstract}
Online Video Large Language Models (VideoLLMs) play a critical role in supporting responsive, real-time interaction. Existing methods focus on streaming perception, lacking a synchronized logical reasoning stream. However, directly applying test-time scaling methods incurs unacceptable response latency. To address this trade-off, we propose \textit{\textbf{V}ideo \textbf{S}treaming \textbf{T}hinking}~(VST), a novel paradigm for streaming video understanding. It supports a \textit{thinking-while-watching} mechanism, which activates reasoning over incoming video clips during streaming. This design improves timely comprehension and coherent cognition while preserving real-time responsiveness by amortizing LLM reasoning latency over video playback. Furthermore, we introduce a comprehensive post-training pipeline that integrates VST-SFT, which structurally adapts the offline VideoLLM to causal streaming reasoning, and VST-RL, which provides end-to-end improvement through self-exploration in a multi-turn video interaction environment. Additionally, we devise an automated training-data synthesis pipeline that uses video knowledge graphs to generate high-quality streaming QA pairs, with an entity–relation grounded streaming Chain-of-Thought to enforce multi-evidence reasoning and sustained attention to the video stream. Extensive evaluations show that VST-7B performs strongly on online benchmarks, \eg $79.5\%$ on StreamingBench and $59.3\%$ on OVO-Bench. Meanwhile, VST remains competitive on offline long-form or reasoning benchmarks. Compared with Video-R1, VST responds $15.7\times$ faster and achieves $+5.4\%$ improvement on VideoHolmes, demonstrating higher efficiency and strong generalization across diverse video understanding tasks. Code, data, and models have been released at \url{https://github.com/1ranGuan/VST}.


\end{abstract}

\section{Introduction}
\begin{figure*}[htbp]
    \centering
    \includegraphics[width=1.0
    \textwidth]{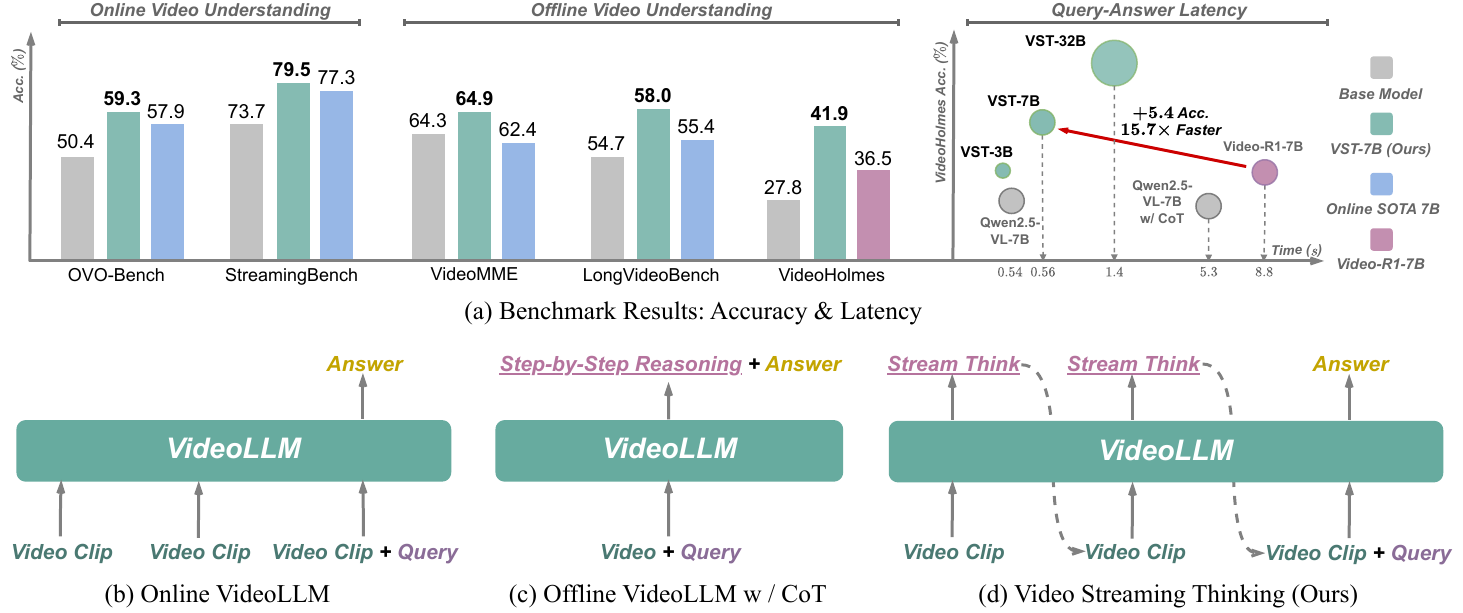}
    \caption{\textbf{Benchmark results and paradigm comparison.} (a) VST-7B delivers strong performance on online and offline video understanding benchmarks while maintaining low QA latency. (b) Existing streaming VideoLLMs focus on efficient streaming processing, but lack explicit analytical reasoning. (c) VideoLLM with CoT performs heavy post-query step-by-step reasoning to improve performance, but incurs high QA latency. (d) Our method introduces proactive pre-query reasoning, interleaving it with video consumption to achieve both strong performance and efficient responsiveness.}
    \vspace{-1.5em}
    \label{fig:fig1}
\end{figure*}

Online video understanding enables Video Large Language Models (VideoLLMs) to interpret streaming visual inputs and respond in real time, making it particularly valuable for embodied intelligence \cite{fu2025minddrive,fang2026towards} and interactive AI assistants \cite{driess2023palm,chen2025livecc}.
Unlike offline methods that benefit from post-hoc global access to the entire video \cite{bai2025qwen2,coreteam2025mimovltechnicalreport,li2024monkey,huang2025mini,jiang2025vknowu}, the core challenges of online video understanding lie in strict temporal causality, real-time processing, and a finite context window. 

Several prior methods have been proposed to address the challenges of online video understanding.
As shown in \cref{fig:fig1}(b), they primarily improve context-window efficiency by explicitly managing visual tokens for compression~\cite{song2024moviechat,yao2025timechat,zengstreamforest,wang2026curvestream} or by retrieving from the KV cache~\cite{distreaming,ning2025livevlm,yang2025streammem}.
However, these methods primarily focus on streaming perception and treat the management of visual features as a form of memory, with limited involvement of the LLM itself and no explicit reasoning or analytical deliberation. 
To fill this missing piece, one promising direction inspired by offline video understanding is to apply test-time scaling via Chain-of-Thought (CoT) to elicit stronger reasoning ability \cite{guo2025deepseek,feng2025video,chenscaling,zeng2026video,zhu2025shuffle,guanthinkomni,liang2025cook,wu2026generation}, as shown in \cref{fig:fig1}(c). Nevertheless, directly performing step-by-step reasoning after the user query can significantly increase QA response latency, making it difficult to meet strict real-time requirements in online scenarios.

In this paper, we introduce the \textit{\textbf{V}ideo \textbf{S}treaming \textbf{T}hinking}~(VST) to resolve the trade-off between explicit reasoning and real-time responsiveness, shifting the LLM backend from passive waiting to active, intermittent reasoning during video consumption.
This design is inspired by insights from human cognition. Findings on \emph{neural coupling}~\cite{hasson2004intersubject,stephens2010speaker} suggest that the logical flow in the brain synchronizes closely with the influx of external information, fostering the perception of current signals and their synthesis into a coherent understanding. 
Similarly, as illustrated in \cref{fig:fig1}(d), our method continuously processes incoming video clips and produces intermediate thoughts in real time. This eliminates the need to defer heavy computation until the query arrives, which is a common limitation of offline VideoLLMs with CoT~\cite{feng2025video,chenscaling,wang2025videorft}.
This \textit{thinking-while-watching} mechanism maintains a coherent internal state over the stream, ensuring that the final response is grounded in a deeply processed understanding of the historical context.
By front-loading and amortizing the reasoning cost ahead of query arrival, VST preserves the low QA latency required in streaming scenarios.

We instantiate this paradigm with a dedicated post-training pipeline that combines supervised fine-tuning (VST-SFT) and reinforcement learning (VST-RL). 
Concretely, we cast streaming thinking as a multi-turn conversation, where the model incrementally writes textual thoughts to an external memory while observing incoming video clips under a constrained visual context window.
In the VST-SFT stage, we align the model with the desired streaming reasoning protocol by learning from off-policy demonstrations that strictly respect temporal causality, thereby bootstrapping its basic \textit{thinking-while-watching} capability. 
Building upon this initialization, the VST-RL stage performs end-to-end reinforcement learning with verifiable rewards, encouraging the model to make intermediate reasoning steps that improve downstream question answering under realistic online conditions. 

Due to the scarcity of existing data for video streaming thinking, we develop an automated synthesis pipeline to support our training, particularly the VST-SFT stage that requires high-quality reasoning demonstrations. Specifically, we model entities and their temporal relationships within long videos as knowledge graphs. By sampling paths from these graphs to form evidence chains, we prompt an offline VideoLLM to generate complex QA pairs and their corresponding intermediate CoTs. This design enforces multi-hop reasoning across diverse visual evidence while ensuring strict alignment between the generated thoughts and the video context. Ultimately, we synthesize a large-scale dataset comprising 100K high-quality streaming reasoning samples.

We conducted extensive evaluations across multiple online and offline video understanding benchmarks (see \cref{fig:fig1}(a)). The results show that our method achieves state-of-the-art performance compared to existing online VideoLLMs, while remaining competitive on offline video understanding benchmarks. Notably, VST performs particularly well on long-form videos that require comprehensive plot comprehension and multi-step reasoning. Moreover, compared to Video-R1, our method achieves higher accuracy while significantly reducing QA latency, demonstrating that VST is a viable test-time scaling approach that meets the requirements of streaming scenarios.

In summary, our main contributions are as follows:

\begin{itemize}[noitemsep, topsep=0pt, parsep=0pt, partopsep=0pt, leftmargin=*]
    \item[$\bullet$] We propose the VST paradigm to interleave active explicit CoT generation with continuous video streams, enabling amortized test-time scaling with real-time responsiveness.

    \item[$\bullet$] A knowledge-graph-based data synthesis pipeline and a dedicated post-training recipe (VST-SFT and VST-RL) are introduced to adapt an offline VideoLLM to streaming settings with strong streaming reasoning capabilities.

    \item[$\bullet$] Extensive evaluations across multiple online and offline video understanding benchmarks demonstrate state-of-the-art performance. In addition, compared to offline CoT VideoLLM, our method provides significantly lower QA latency.
\end{itemize}

\label{intro}

\section{Method}
\label{method} 
\begin{figure*}[tbp]
    \centering
    \includegraphics[width=1.0\textwidth]{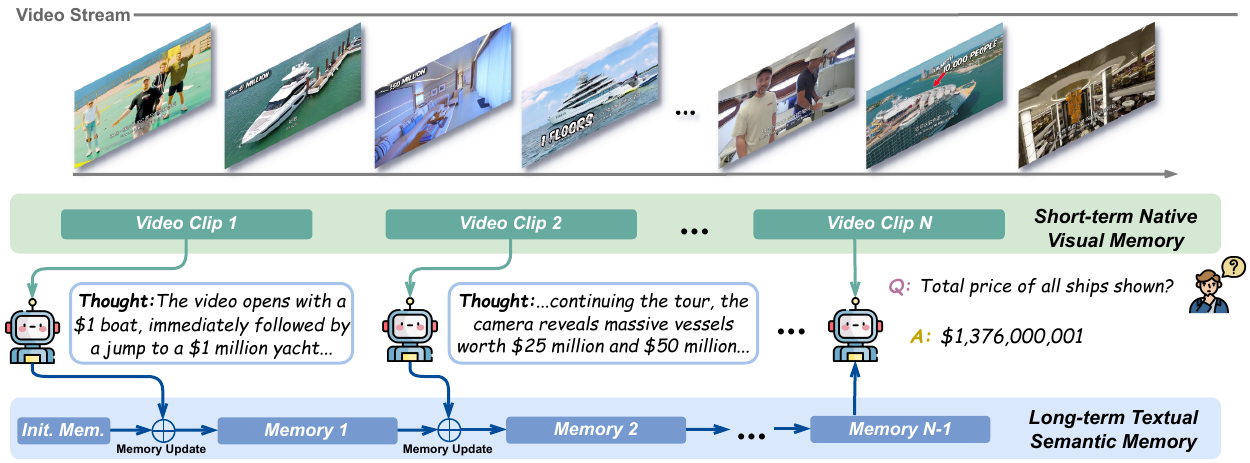}
    \caption{\textbf{Illustration of the Video Streaming Thinking pipeline.} The model employs a streaming thought mechanism to compress visual dynamics into a long-term textual memory. Combined with the short-term visual buffer, this enables efficient reasoning over indefinite video streams with fixed memory budgets.}
    \label{fig:vst_pipeline}
\end{figure*}

\subsection{The \textit{\textbf{V}ideo \textbf{S}treaming \textbf{T}hinking}~(VST) Paradigm}
\label{sec: paradigm}
We formulate VST as a multi-round video conversation task operating within a constrained context window, as illustrated in \cref{fig:vst_pipeline}.
Unlike previous online VideoLLMs, our model leverages streaming intervals before a user query to proactively reason about the content via autoregressive textual generation. This process synthesizes key visual details and event dynamics into a dual-memory system: maintaining a \textit{short-term native video memory} for the current visual context, while accumulating a \textit{long-term textual semantic memory} of past events.

Formally, given a video stream $\mathcal{V}$, let $\mathbf{v}_i$ denote the visual features for the $i$-th frame. We accumulate these incoming features into discrete clips $\mathbf{c}^k = \{\mathbf{v}_i\}_{i=\tau_{k-1}+1}^{\tau_k}$, where the boundary $\tau_k$ is set when the accumulated visual tokens reach the preset capacity $L$. At each interval $k$, conditioned on the current clip $\mathbf{c}^k$ and the accumulated memory $\mathbf{m}^{k-1}$, the LLM generates a streaming thought $\mathbf{z}^k$ by sampling from the distribution $\mathbf{z}^k \sim p(\mathbf{z} \mid \mathbf{c}^k, \mathbf{m}^{k-1})$. Here, $\mathbf{z}^k$ summarizes the essential semantics of the current video segment, preserving the continuity of the overall thought process. For the long-term textual memory, we employ a memory update function $\mathbf{m}^k = \text{Update}(\mathbf{m}^{k-1}, \mathbf{z}^k)$, which adopts a simple first-in-first-out strategy to evict the earliest memory entries.

This iterative reasoning process continues until step $K$, when a user query $\mathbf{q}$ is received. Upon this trigger, the LLM generates the final response $\mathbf{y}$ based on the accumulated thoughts $\{\mathbf{z}^k\}_{k=1}^{K-1}$ and the latest visual context $\mathbf{c}^K$. Consequently, the joint probability of the thoughts and the final answer is decomposed as:
{
\begin{equation} \label{eq: general reasoning}
p\!\left(\mathbf{y}, \{\mathbf{z}^k\}_{k=1}^{K-1} \,\middle|\, \mathbf{q}, \mathcal{V}\right) = \underbrace{p(\mathbf{y} \mid \mathbf{q}, \mathbf{c}^K, \mathbf{m}^{K-1})}_{\text{Direct Answer}} \prod_{k=1}^{K-1} \underbrace{p(\mathbf{z}^k \mid \mathbf{c}^k, \mathbf{m}^{k-1})}_{\text{Streaming Thinking}}.
\end{equation}
}%
This formulation yields two distinct advantages. 1) It amortizes the computational cost of Chain-of-Thought (CoT) generation over the pre-query phase. This strategy effectively achieves test-time scaling to boost performance without incurring additional latency at the moment of user interaction. 2) The sequential generation of thoughts naturally aligns with the temporal causality inherent in streaming videos. This structure facilitates the adaptation of offline models to online scenarios by mirroring the progressive nature of the video stream.

\subsection{Training Method for VST}
To instantiate the VST paradigm introduced in \cref{sec: paradigm}, we develop a two-stage post-training pipeline that combines supervised fine-tuning (VST-SFT) and reinforcement learning (VST-RL), progressively endowing an offline VideoLLM with streaming thinking capabilities.
The VST-SFT stage adapts the offline model to the temporal causality of streaming video, while learning reasoning capabilities from off-policy expert data. 
Subsequently, VST-RL transitions the model from off-policy imitation to on-policy RL, and refines these learned capabilities for further end-to-end improvement.

\begin{figure*}[tbp]
    \centering
    \includegraphics[width=1.0\textwidth]{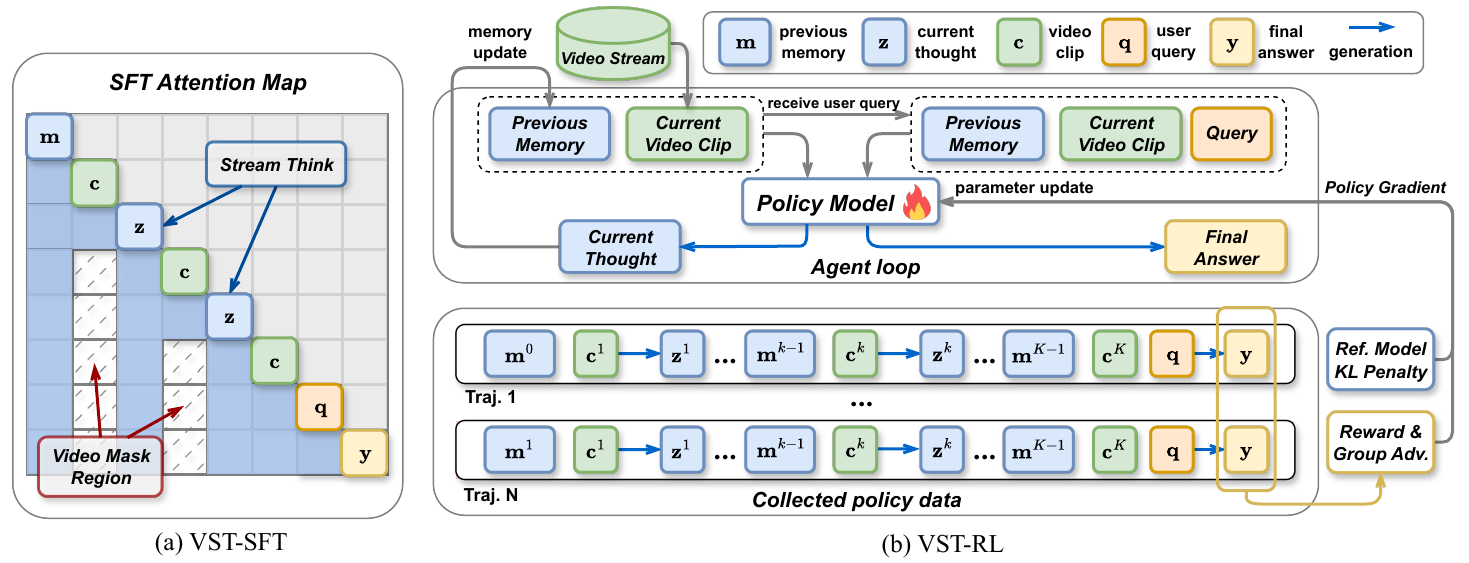}
    \caption{\textbf{Overview of the training pipeline.} (a) VST-SFT applies a streaming attention mask to enforce temporal causality, restricting attention to the current visual buffer and history textual context. (b) VST-RL performs on-policy optimization via an agentic loop, improving the quality of streaming thoughts through verifiable rewards computed solely from the final answer.}
    \label{fig:vst_training}
\end{figure*}

\subsubsection{Stage 1: VST-SFT.}
We initiate the training pipeline with SFT to instill the streaming thought mechanism into the offline VideoLLM. 
For a training instance, we explicitly formulate the sequence as:
{
\begin{equation}  \label{eq: sft data}
    \mathcal{S} = \Big( \mathbf{m}^0, (\mathbf{c}^1, \mathbf{z}^1), \dots, (\mathbf{c}^{K-1}, \mathbf{z}^{K-1}), \mathbf{c}^K, \mathbf{q}, \mathbf{y} \Big).
\end{equation}
}%
Here, $\mathbf{m}^0$ denotes the initial memory, and $(\mathbf{c}^k, \mathbf{z}^k)$ represent the interleaved video clips and streaming thoughts. The sequence concludes with the final clip $\mathbf{c}^K$, user query $\mathbf{q}$, and ground truth response $\mathbf{y}$.

To align with the streaming inference architecture, we apply a streaming video attention mask.
As depicted in \cref{fig:vst_training}(a), this mask restricts the model's attention to a fixed-size window of recent visual tokens, mirroring the short-term visual buffer used during inference.
Specifically, let $M$ be the additive attention mask.
Let $\mathbb{I}_{v}(j)\in\{0,1\}$ indicate whether the $j$-th token is a visual token, and let $L$ denote the visual buffer size.
Therefore, the attention mask can be written as:
\begin{equation}
    M_{i,j}=
    \begin{cases}
    0, & j \le i \ \text{and}\ \Big(\mathbb{I}_{v}(j)=0 \ \text{or}\ \sum\nolimits_{t=j+1}^{i}\mathbb{I}_{v}(t) < L \Big)\\
    -\infty, & \text{otherwise}
    \end{cases}
    \label{eq:streaming_video_mask}
\end{equation}
In this way, the model can only access a sliding window of the latest $L$ visual tokens, while all non-visual tokens remain fully visible under the causal constraint.
Furthermore, to accommodate context length constraints while handling long-form videos, we implement a temporal segmentation strategy. 
The original sequence $\mathcal{S}$ is sliced into consecutive segments $\{\mathbf{s}_n\}_{n=1}^M$, defined as:
\begin{equation}
    \mathbf{s}_n =
    \begin{cases}
    \Big(\mathbf{m}^{n-1}, \{(\mathbf{c}^k, \mathbf{z}^k)\}_{k=T_{n-1}+1}^{T_n}\Big), & n < M \\[3pt]
    \Big(\mathbf{m}^{n-1}, \{(\mathbf{c}^k, \mathbf{z}^k)\}_{k=T_{n-1}+1}^{K-1}, \mathbf{c}^{K}, \mathbf{q}, \mathbf{y}\Big), & n = M
    \end{cases}
    \label{eq:video_clip}
\end{equation}
where $T_n$ denotes the cut-off index for the $n$-th segment. The memory state is updated recursively across segments following $\mathbf{m}^n = \text{Update}(\mathbf{m}^{n-1}, \{\mathbf{z}^k\}_{k=T_{n-1}+1}^{T_{n}})$. 
During SFT, we apply the standard next-token prediction loss exclusively to the streaming thoughts $\{\mathbf{z}^k\}_{k=1}^{K-1}$ and the final response $\mathbf{y}$, treating visual tokens and historical memory as conditioning inputs. 

\subsubsection{Stage 2: VST-RL.}
Building upon the supervised foundation, we introduce VST-RL to transition the model from off-policy imitation to on-policy self-improvement. 
The RL training process consists of two main phases: trajectory rollout and policy gradient optimization.

As shown in the upper part of \cref{fig:vst_training}(b), the rollout phase operates as an agentic loop. 
The policy model interacts with the streaming environment to generate a trajectory $\mathcal{T}$ following the predefined joint probability in \cref{eq: general reasoning}, where the streaming thoughts $\mathbf{\hat{z}}^k$ and the final response $\mathbf{\hat{y}}$ are sequentially sampled from the sampling policy $\pi_{\theta'}$.
After collecting a group of $N$ trajectories $\{\mathcal{T}_i\}_{i=1}^{N}$, we employ a GRPO \cite{guo2025deepseek,yu2025dapo,yu2025memagent,liu2025understanding} strategy to optimize the policy model. 
We compute the reward $\mathbf{r_i}$ solely based on the final answer $\mathbf{y_i}$ via verifiable reward functions. 
To encourage the model to generate useful streaming thoughts, the calculated advantage is assigned to all generated tokens within the entire trajectory $\mathcal{T}_i$. The policy gradient objective is calculated as:
{\small
\begin{equation}
\label{eq: policy_gradient}
\mathcal{J}_{\text{RL}}(\theta) = \mathbb{E}_{q\sim \mathcal{D},\{\mathcal{T}_i\}_{i=1}^{N} \sim \pi_{\theta'}(\cdot \mid q)} \left[ \frac{1}{\sum_{i=1}^N|\mathcal{T}_i|} \sum_{i=1}^N \sum_{t=1}^{|\mathcal{T}_i|} \left( \mathcal{L}^{\text{clip}}_{i,t}(\theta) - \beta D_\text{KL}(\pi_{\theta}||\pi_{\text{ref}}) \right) \right],
\end{equation}
}%
\begin{equation}
\label{eq: clip_term}
\mathcal{L}^{\text{clip}}_{i,t}(\theta) = \min \left[ \gamma_t(\theta) \hat{A}_{i}, \text{clip} \left( \gamma_t(\theta), 1 - \epsilon_\text{low}, 1 + \epsilon_\text{high} \right) \hat{A}_{i} \right].
\end{equation}
Where $|\mathcal{T}_i|$ denotes the total number of generated tokens in trajectory $\mathcal{T}_i$, $\gamma_t(\theta)$ represents the probability ratio between $\pi_\theta$ and the sampling policy $\pi_{\theta'}$ at step $t$, $\hat{A}_i = r_i - \text{mean}(R)$ is the group relative advantage, and $\epsilon_{\text{low}}, \epsilon_{\text{high}}$ are the clipping hyperparameters follow DAPO \cite{yu2025dapo}.

\begin{figure}[t]
    \centering
    \includegraphics[width=\linewidth]{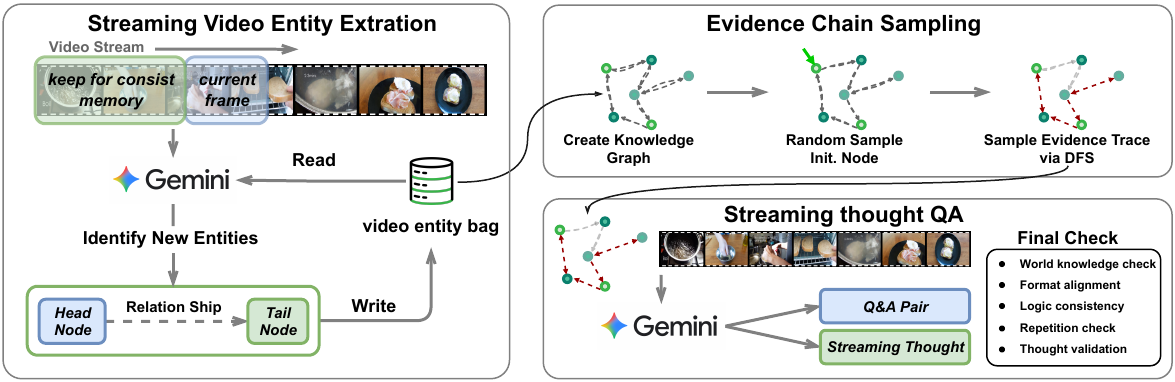} % Placeholder for the uploaded image
    \caption{\textbf{Stream-Thought QA data curation pipeline.} We incrementally extract video entities and relations to build a knowledge graph, sample multi-hop evidence chains, and use Gemini to generate streaming QA pairs with grounded streaming thoughts, followed by automatic filtering.}
    \label{fig:pipeline}
\end{figure}

\subsection{Data Synthesis Pipeline for VST}
\label{subsec:data_curation}

We generate a set of video streaming thought data to support VST training, motivated by the fact that most existing chain-of-thought (CoT) datasets target offline VideoLLMs with a global, hindsight view of the entire video, making it difficult to avoid information leakage under causal streaming constraints. To this end, we introduce an automated data generation pipeline grounded in knowledge graphs. As illustrated in \cref{fig:pipeline}, the pipeline produces high-quality training examples with explicit reasoning paths through streaming video entity extraction, evidence chain sampling, and streaming thought QA synthesis.

\subsubsection{Streaming Video Entity Extraction.}
To build a temporally consistent knowledge graph, we maintain an \emph{entity bank} and extract triples from a sliding window over the video stream. We segment the video into $N$ scene clips with PySceneDetect. For each incoming clip, an offline VideoLLM (\eg Gemini~3.0 flash) updates the entity bank by adding newly observed entities and relations as $\big(\textbf{head},\, \text{relation},\, \textbf{tail}\big)$. When the window exceeds size $W$, we drop the oldest clip and retain the most recent $W-1$ overlapping clips to preserve temporal continuity. The entity bank thus serves as a lightweight memory for consistent entity tracking and timeline-aligned graph construction.

\subsubsection{Evidence Chain Sampling.}
After processing the whole video, the complete entity bank is refined using an LLM to filter out noise entities, such as duplicates and subtitles. Subsequently, NetworkX \cite{hagberg2007exploring} is used to construct the knowledge graph, which represents the logical relationships between events in the video. To mine long-term causal dependencies, an initial node is randomly selected, and a depth-first search (DFS) is used to extract evidence chains. Each node in these chains contains detailed information about the head and tail entities, their relationship, timestamps, and scene descriptions, facilitating comprehensive reasoning over the video content. For each video, we sample multiple evidence chains, enforcing that the entity overlap between any two chains is below $10\%$ to promote diversity.

\subsubsection{Stream Thought QA Synthesis.}
The final phase leverages Gemini~3.0 flash as a data synthesizer. Conditioned on the video knowledge graph, the model first generates a streaming CoT rationale to actively reason over video events and dynamic content. Subsequently, aligned with a sampled evidence chain $\{\mathbf{z}^k\}_{k=1}^{K}$, it synthesizes a query $\textbf{q}$ and the final answer $\textbf{y}$, necessitating multi-evidence reasoning that integrates the CoT with visual context. To ensure data fidelity, we apply a strict post-generation filtering rubric, including: world-knowledge check, format alignment, logical consistency, repetition check, and thought validation.

\subsubsection{Curation of VST training set.}
Following the above procedure, we generate 100K streaming-thought examples with videos from LLaVA-Vid\cite{zhang2025llavavideo} and Video-Marathon\cite{lin2025unleashing}.
In addition, our full supervised fine-tuning corpus for VST-SFT includes 50K open-ended QA instances randomly sampled from LLaVA-Vid. For VST-RL, we train on 11K sampled questions, including multiple-choice questions from LLaVA-Vid, Video-Marathon, and Onethinker~\cite{feng2025onethinker}, as well as counting questions from RepCount~\cite{hu2022transrac}.

\section{Experiment}
\label{sec:exp}
\subsection{Implementation Details}
We adopt Qwen2.5-VL~\cite{bai2025qwen2} as our base offline VideoLLM, processing input videos at 2 fps. Both VST-SFT and VST-RL (7B model) training stages are conducted on 32 $\times$ 80GB VRAM GPUs, utilizing the datasets detailed in \cref{subsec:data_curation}. The visual encoder and projection layer are frozen throughout the entire training process. For VST-SFT, each training sample follows a 128 second time limit, and overlong raw videos are segmented into clips following \cref{eq:video_clip}. For VST-RL, we employ verl~\cite{sheng2024hybridflow} with vLLM~\cite{kwon2023efficient} and FSDP~\cite{zhao2023pytorch} backend. We configure the rollout batch size to 256 with a group size of $N=8$, and define the reward function based on the correctness of the final answer. Additionally, following LongVILA-R1~\cite{chenscaling}, we leverage the paralleled encoding strategy during rollout to pre-compute video embeddings. During testing, following StreamingForest~\cite{zengstreamforest}, we cap each inference step (including streaming-think and the final answer) at $8{,}192$ video tokens and limit the max thinking times to $4$ for efficient evaluation. We conduct all evaluations using the lmms-eval framework~\cite{zhang2024lmmsevalrealitycheckevaluation}.

%%%%%%%%%%%%%%%%%%%%%%%%%%%
% streamingbench
%%%%%%%%%%%%%%%%%%%%%%%%%%%
\begin{table*}[!ht]
\caption{Comparison of offline and online VideoLLMs on StreamingBench Real-Time understanding tasks.}
\centering
% \resizebox{\textwidth}{!}{%
\fontsize{7pt}{8pt}\selectfont
\setlength{\tabcolsep}{2pt}
\begin{tabular}{l c| c c c c c c c c c c |c}
\toprule
Model & Venue & OP & CR & CS & ATP & EU & TR & PR & SU & ACP & CT & \textbf{Overall} \\
\midrule
\multicolumn{13}{l}{\textit{Proprietary Models}} \\
\midrule
Gemini 1.5 pro \cite{team2024gemini} & - & 79.0 & 80.5 & 83.5 & 79.7 & 80.0 & 84.7 & 77.8 & 64.2 & 72.0 & 48.7 & 75.7 \\
GPT-4o \cite{gpt4o} & - & 77.1 & 80.5 & 83.9 & 76.5 & 70.2 & 83.8 & 66.7 & 62.2 & 69.1 & 49.2 & 73.3 \\
\midrule
\multicolumn{13}{l}{\textit{Open-source Offline Models}} \\
\midrule
% Video-LLaMA2-7B & 7B & 55.9 & 55.5 & 57.4 & 58.2 & 52.8 & 43.6 & 39.8 & 42.7 & 45.6 & 35.2 & 49.5 \\
VILA-1.5-8B\cite{lin2024vila} & CVPR'24 & 53.7 & 49.2 & 71.0 & 56.9 & 53.4 & 53.9 & 54.6 & 48.8 & 50.1 & 17.6 & 52.3 \\
LongVA-7B\cite{zhang2025long} & TMLR'25 & 70.0 & 63.3 & 61.2 & 70.9 & 62.7 & 59.5 & 61.1 & 53.7 & 54.7 & 34.7 & 60.0 \\
MiniCPM-v2.6-7B\cite{hu2024minicpm} & COLM'24 & 71.9 & 71.1 & 77.9 & 75.8 & 64.6 & 65.7 & 70.4 & 56.1 & 62.3 & 53.4 & 67.4 \\
LLaVA-OV-7B\cite{li2025llavaonevision} & TMLR'25 & 80.4 & 74.2 & 76.0 & 80.7 & 72.7 & 71.7 & 67.6 & 65.5 & 65.7 & 45.1 & 71.1 \\
Qwen2.5-VL-7B\cite{bai2025qwen2} & - & 78.3 & 80.5 & 78.9 & 80.5 & 76.7 & 78.5 & 79.6 & 63.4 & 66.2 & 53.2 & 73.7 \\
\midrule
\multicolumn{13}{l}{\textit{Open-source Online Models}} \\
\midrule
Flash-VStream-7B\cite{zhang2025flash} & ICCV'25 & 25.9 & 43.6 & 24.9 & 23.9 & 27.3 & 13.1 & 18.5 & 25.2 & 23.9 & 48.7 & 23.2 \\
VideoLLM-online-8B\cite{chen2024videollm} & CVPR'24 & 39.1 & 40.1 & 34.5 & 31.1 & 46.0 & 32.4 & 31.5 & 34.2 & 42.5 & 27.9 & 36.0 \\
Dispider-8B\cite{qian2025dispider} & CVPR'25 & 74.9 & 75.5 & 74.1 & 73.1 & 74.4 & 59.9 & 76.1 & 62.9 & 62.2 & 45.8 & 67.6 \\
TimeChatOnline-7B\cite{yao2025timechat} & MM'25 & 80.2 & 82.0 & 79.5 & 83.3 & 76.1 & 78.5 & 78.7 & 64.6 & 69.6 & 58.0 & 75.4 \\
Streamforest-7B\cite{zengstreamforest}& NeurIPS'25 & 83.1 & 82.8 & 82.7 & 84.3 & 77.5 & 78.2 & 76.9 & 69.1 & 75.6 & 54.4 & 77.3 \\
\midrule
\rowcolor{cyan!15} \textbf{VST-7B (ours)} & ECCV'26 & 85.4 & 82.0 & 86.4 & 89.1 & 74.2 & 87.2 & 82.4 & 73.1 & 73.9 & 47.3 & 79.5\\
\bottomrule
\end{tabular}%
% }
\label{tab:streamingbench}
\end{table*}

\subsection{Benchmarks}
To demonstrate the effectiveness of our method, we conducted a comprehensive evaluation across five video understanding benchmarks. Specifically, StreamingBench~\cite{lin2024streamingbench} and OVO-Bench~\cite{niu2025ovo} are utilized for online video understanding, focusing on the model's online reasoning capabilities and temporal awareness. VideoMME~\cite{fu2025video} serves as a comprehensive offline benchmark covering diverse domains and varying video durations. LongVideoBench~\cite{wu2024longvideobench} is designed to evaluate the long-form video understanding capabilities, while Video-Holmes~\cite{cheng2025video} emphasizes logical reasoning within video content.

\subsection{Online Video Benchmark Results}
As shown in \cref{tab:streamingbench,tab:ovobench}, we evaluate our model on StreamingBench and OVO-Bench. VST-7B achieves $79.5\%$ on StreamingBench and $59.3\%$ on OVO-Bench, clearly outperforming prior open-source streaming SOTA models, including Streamforest \cite{zengstreamforest} ($77.3\%$) on StreamingBench and Streamo \cite{xia2025streaming} ($57.9\%$) on OVO-Bench.
Notably, despite being much smaller than proprietary models, our method surpasses GPT-4o and Gemini 1.5 pro on StreamingBench by $6.2\%$ and $3.8\%$, respectively, and achieves comparable performance with GPT-4o on OVO-Bench.
Beyond the overall scores, VST-7B is particularly strong on OVO-Bench’s Backward Tracing task, where it achieves 
56.7\%, outperforming Streamforest by +4.7\%. This result indicates that our model can retain and retrieve historical information effectively, supporting sustained memory over streaming inputs.
These results highlight the strength of our approach for streaming video understanding. We believe the gains stem from our VST paradigm and a tailored post-training recipe, which together improve the model’s ability.

\begin{table*}[!ht]
\caption{Comparison of offline and online VideoLLMs on OVO-Bench.}
\centering
\resizebox{\textwidth}{!}{%
\begin{tabular}{l c |cccccc|c| ccc|c| ccc|c| c}
\toprule
\multirow{2}{*}{Model} & \multirow{2}{*}{Venue} & \multicolumn{7}{c}{Real-Time} & \multicolumn{4}{c}{Backward} & \multicolumn{4}{c}{Forward} & \textbf{Overall}  \\
\cmidrule(lr){3-9} \cmidrule(lr){10-13} \cmidrule(lr){14-17} \cmidrule(lr){18-18}
 & & OCR & ACR & ATR & STU & FPD & OJR & Avg. & EPM & ASI & HLD & Avg. & REC & SSR & CRR & Avg. & Avg. \\
\midrule
\multicolumn{18}{l}{\textit{Proprietary Models}} \\
\midrule
Gemini 1.5 pro \cite{team2024gemini} & - & 85.9 & 67.0 & 79.3 & 58.4 & 63.4 & 62.0 & 69.3 & 58.6 & 76.4 & 52.6 & 62.5 & 35.5 & 74.2 & 61.7 & 57.2 & 63.0 \\
GPT-4o \cite{gpt4o} & - & 69.8 & 64.2 & 71.6 & 51.1 & 70.3 & 59.8 & 64.5 & 57.9 & 75.7 & 48.7 & 60.8 & 27.6 & 73.2 & 59.4 & 53.4 & 59.5 \\
\midrule
\multicolumn{18}{l}{\textit{Open-source Offline Models}} \\
\midrule
Qwen2-VL-72B\cite{wang2024qwen2} & - & 65.8 & 60.6 & 69.8 & 51.7 & 69.3 & 54.4 & 61.9 & 52.5 & 60.8 & 57.5 & 57.0 & 38.8 & 64.1 & 45.0 & 49.3 & 56.3 \\
LLaVA-Video-7B\cite{zhang2025llavavideo} & TMLR'25 & 69.1 & 58.7 & 68.8 & 49.4 & 74.3 & 59.8 & 63.5 & 56.2 & 57.4 & 7.5 & 40.4 & 34.1 & 70.0 & 60.4 & 54.8 & 52.9 \\
LLaVA-OV-7B\cite{li2025llavaonevision} & TMLR'25 & 66.4 & 57.8 & 73.3 & 53.4 & 71.3 & 62.0 & 64.0 & 54.2 & 55.4 & 21.5 & 43.7 & 25.6 & 67.1 & 58.8 & 50.5 & 52.7 \\
LongVU-7B\cite{shenlongvu} & ICML'25 & 53.7 & 53.2 & 62.9 & 47.8 & 68.3 & 59.8 & 57.6 & 40.7 & 59.5 & 4.8 & 35.0 & 12.2 & 69.5 & 60.8 & 47.5 & 46.7 \\
% Qwen2.5-VL-7B &&&&&&&&63.4&&&& 46.4 &&&&41.5&50.4\\
\midrule
\multicolumn{18}{l}{\textit{Open-source Online Models}} \\
\midrule
VideoLLM-online-8B\cite{chen2024videollm} & CVPR'24 & 8.1 & 23.9 & 12.1 & 14.0 & 45.5 & 21.2 & 20.8 & 22.2 & 18.8 & 12.2 & 17.7 & - & - & - & - & - \\
Dispider-8B\cite{qian2025dispider} & CVPR'25 & 57.7 & 49.5 & 62.1 & 44.9 & 61.4 & 51.6 & 54.6 & 48.5 & 55.4 & 4.3 & 36.1 & 18.1 & 37.4 & 48.8 & 34.7 & 41.8 \\
TimeChatOnline-7B\cite{yao2025timechat} & MM'25 &75.2 &46.8 &70.7 &47.8 &69.3 &61.4 &61.9& 55.9 &59.5 &9.7 &41.7 &31.6 &38.5 &40.0& 36.7& 46.7\\
Streamforest-7B\cite{zengstreamforest}& NeurIPS'25 &68.5&53.2&71.6&47.8&65.4&60.9&61.2&58.9&64.9&32.3&52.0&32.8&70.6&57.1&52.5&55.6 \\
Streamo-7B\cite{xia2025streaming} & CVPR'26 &77.2& 66.1 &76.7 &45.5 &66.3 &72.8 &67.4 &55.6 &58.1 &33.9 &49.2 &30.8 &57.6 &82.5& 57.0 &57.9\\
\midrule 
\rowcolor{cyan!15} \textbf{VST-7B (ours)}  & ECCV'26 & 80.5 & 55.1 &72.4&55.1&76.2&64.1&67.2&56.9&64.9&48.4&56.7&33.0&66.9&62.1&54.0&59.3\\
\bottomrule
\end{tabular}%
}
\label{tab:ovobench}
\end{table*}
%%%%%%%%%%%%%%%%%%%%%%%%%%%
% Offline Benchmarks
%%%%%%%%%%%%%%%%%%%%%%%%%%%
\begin{table*}[!ht]
\centering
\caption{Comparison of offline and online VideoLLMs on VideoMME (without subtitles), LongVideoBench, and VideoHolmes.}
\fontsize{6pt}{7pt}\selectfont
\setlength{\tabcolsep}{5pt}
\begin{tabular}{lc|cccc}
\toprule
\multirow{2}{*}{Model} & \multirow{2}{*}{Venue} & \multicolumn{2}{c}{\textbf{VideoMME} w/o sub.} & \multirow{2}{*}{\textbf{LongVideoBench}} & \multirow{2}{*}{\textbf{VideoHolmes}}\\
\cmidrule(lr){3-4}
 & & Long & Overall & \\
\midrule
\multicolumn{6}{l}{\textit{Proprietary Models}} \\
\midrule
Gemini 1.5 pro \cite{team2024gemini} & - & 67.4 & 75.0 & 64.0 & 45.7\\
GPT-4o \cite{gpt4o}         & - & 65.3 & 71.9 & 66.7 & 42.0\\
\midrule
\multicolumn{6}{l}{\textit{Open-source Offline Models}} \\
\midrule
LongVA-7B \cite{zhang2025long}       & TMLR’25 & 47.6 & 54.3 & 56.3 & -\\
Video-R1-7B \cite{feng2025video}    & NeurIPS'25 & - & 61.4 & - & 36.5\\
LongVILA-R1-7B \cite{chenscaling} & NeurIPS'25 & 55.2 & 65.1 & 58.0 & - \\
REVISOR-7B \cite{li2025revisor}     & CVPR'26    & 56.2 & 65.7 & 57.5 & - \\
\midrule
\multicolumn{6}{l}{\textit{Open-source Online Models}} \\
\midrule
Dispider-7B \cite{qian2025dispider}        & CVPR'25 & - & 57.2 & - & -\\
Streamforest-7B \cite{zengstreamforest}    & NeurIPS'25 & - & 61.4 & - & -\\
TimeChatOnline-7B \cite{yao2025timechat}  & MM'25  & 48.4 & 62.4 & 55.4 & - \\
\midrule
\rowcolor{cyan!15} 
\textbf{VST-7B (Ours)} & ECCV'26 & 55.3 & 64.9 & 58.0 & 41.9\\

\bottomrule
\end{tabular}
\label{tab:offline}
\end{table*}

\subsection{Offline Video Benchmark Results}
In \cref{tab:offline}, we evaluate VST-7B on three offline video benchmarks, including VideoMME, LongVideoBench, and VideoHolmes. The results show that VST-7B delivers competitive performance across all three datasets, with particularly strong gains on long-video understanding and complex reasoning. On long-video benchmarks, VST-7B achieves 55.3\% on VideoMME-long, outperforming TimeChat-Online by +6.9\%, and 58.0\% on LongVideoBench, exceeding it by +2.6\%. On the reasoning benchmark VideoHolmes, VST-7B reaches 41.9\%, surpassing Video-R1 by +5.4\%. We attribute these improvements to our streaming thinking framework, which enables dynamic thinking over long videos to build long-term memory, and leverages both historical memory and current visual context for deep reasoning.

%%%%%%%%%%%%%%%%%%%%%%%%%%%
% Ablation 1
%%%%%%%%%%%%%%%%%%%%%%%%%%%
\begin{table*}[t]
\centering
\caption{Ablation study on VST training schedule.}
\fontsize{8pt}{9pt}\selectfont
\setlength{\tabcolsep}{5pt}

\begin{tabular*}{\textwidth}{@{\extracolsep{\fill}} l c c c c @{}}
\toprule
\multirow{2}{*}{Model \& Config}
& \multicolumn{3}{c}{\textbf{OVO-Bench}}
& \textbf{VideoMME} \\
\cmidrule(lr){2-4}
& \scriptsize{Backward}
& \scriptsize{Forward}
& \scriptsize{Overall}
& \scriptsize{w/o sub. Overall} \\
\midrule
Qwen2.5-VL-7B (Base model) & 47.5 & 41.9 & 50.5 & 62.9\\
\midrule
\multicolumn{5}{l}{\scriptsize{\textit{Ablation on VST-SFT training data}}} \\
\midrule
+LLava-Vid (50K) & 49.9 & 42.4 & 52.3 & 61.8\\
+LLava-Vid (30K) \& VST (20K) & 52.0 & 50.1 & 56.8 & 62.5\\
+LLava-Vid (20K) \& VST (30K) & 53.3 & 50.0 & 57.1 & 63.1\\
\midrule
\multicolumn{5}{l}{\scriptsize{\textit{Ablation on different training stage}}} \\
\midrule
+VST-SFT & 56.7 & 48.5 & 57.4 & 63.0\\
+VST-RL  & 49.3 & 54.6 & 56.8 & 62.8\\
+VST-SFT \& VST-RL & 56.7 & 54.0 & 59.3 & 64.9\\
\bottomrule
\end{tabular*}

\label{tab:comparison}
\end{table*}

\begin{figure}[b]
    \centering
    \includegraphics[width=\linewidth]{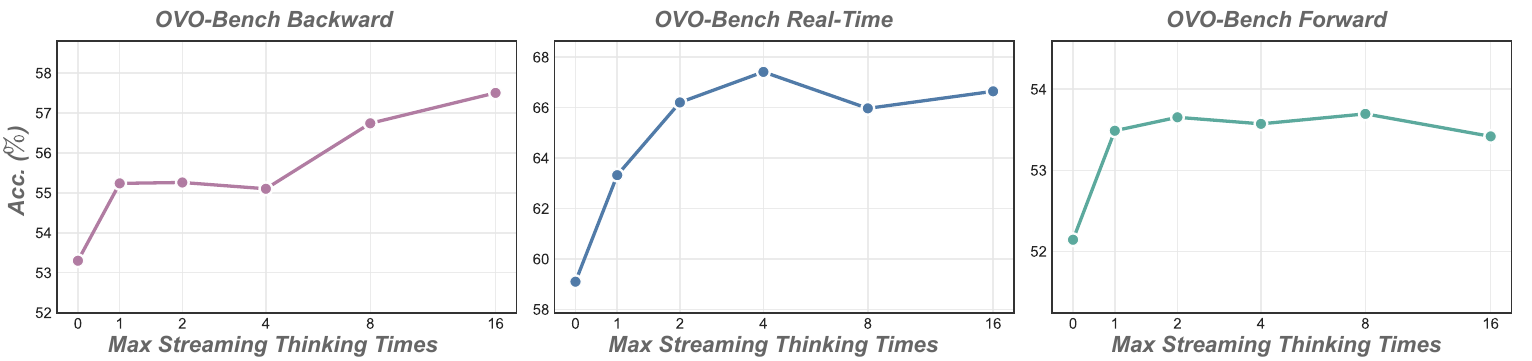} % Placeholder for the uploaded image
    \caption{\textbf{Ablation study on max thinking times.}}
    \label{fig:abl_2}
\end{figure}

\subsection{Ablation Study}

\subsubsection{Ablation on training schedule.}
As shown in \cref{tab:comparison}, we first analyze the composition of the SFT training data. Mixing our VST data with the LLaVA-Vid QA dataset consistently improves online video understanding. Specifically, compared to using 50K LLaVA-Vid data alone, the mix of 20K LLaVA-Vid and 30K VST data achieves a +4.8\% gain on the OVO-Bench.  
Furthermore, the ablation on different training stages demonstrates that our training strategies effectively enhance online video capabilities. Interestingly, we find that VST-SFT primarily benefits the model's backward memory capacity (+9.2\% improvement on Backward track), while VST-RL is advantageous for forward prediction capabilities (improving the Forward score of +12.7\%). Finally, combining both stages (VST-SFT \& VST-RL) yields the highest overall performance on both OVO-Bench (59.3\%) and VideoMME (64.9\%). 

\subsubsection{Ablation on Streaming Thinking Times at Inference.}
\cref{fig:abl_2} analyzes the impact of maximum streaming thinking times on OVO-Bench. For the Backward task, accuracy increases from 53.3\%  and grows continuously from 1 to 16 steps, ultimately reaching 57.5\%. This demonstrates that additional thinking steps help generate precise memories for backward tracing. For the Real-Time and Forward tasks, initial thinking steps clearly aid in understanding visual information. However, performance reaches a plateau for $\ge 4$ steps, as excessive memory details introduce redundancy.

%%%%%%%%%%%%%%%%%%%%%%%%%%%
% Ablation 2
%%%%%%%%%%%%%%%%%%%%%%%%%%%
\begin{table*}[t]
\centering
\caption{Ablation study on the scale of the base offline VideoLLM. All results in this table are obtained with the VST inference pipeline, where Qwen is evaluated as a zero-shot inference baseline.}
\fontsize{8pt}{9pt}\selectfont
\setlength{\tabcolsep}{2pt} 
\begin{tabular}{l l c c c c c}
\toprule
Size & Model & \makecell{\textbf{OVOB.}\\\scriptsize{Overall}} & \makecell{\textbf{StreamingB.}\\\scriptsize{Realtime}} & \makecell{\textbf{V-MME}\\\scriptsize{w/o sub. Overall}} & \textbf{LongVideoB.} & \textbf{VideoHolmes}\\
\midrule
\multirow{2}*{3B} & Qwen2.5-VL&53.1&67.8&57.9&53.3&30.7\\[3pt]
& VST &56.2&75.5&59.5&54.1&36.1\\
\midrule
\multirow{2}*{7B} & Qwen2.5-VL&55.0&71.7&62.3&54.7&32.9\\[3pt]
& VST &59.3&79.5&64.9&58.0&41.9\\
\midrule
\multirow{2}*{32B} & Qwen2.5-VL&60.1&71.5&65.8&59.8&40.1\\[3pt]
& VST &63.5&80.7&67.2&60.7&45.1\\
\bottomrule
\end{tabular}%
% }
\label{tab:abla_model_size}
\end{table*}

\subsubsection{Ablation on Base Model Size.}
\cref{tab:abla_model_size} examines the impact of the base model capacity. Note that the base Qwen rows here are evaluated under the VST streaming inference pipeline (zero-shot), and thus differ from the native offline base in \cref{tab:comparison}; this isolates the contribution of training from that of the inference protocol. We apply our two-stage training recipe (VST-SFT and VST-RL) to the Qwen2.5-VL-Instruct models at 3B, 7B, and 32B scales. The Video Streaming Thinking paradigm yields consistent improvements across all online and offline benchmarks regardless of the model size. For instance, on StreamingBench, VST achieves absolute accuracy gains of +7.7\%, +7.8\%, and +9.2\% over the 3B, 7B, and 32B base models, respectively. Similar consistent enhancements are observed on complex tasks like VideoHolmes (+5.4\%, +9.0\%, and +5.0\%). These results demonstrate that our proposed method is highly parameter-scalable.

\subsection{Analysis}
\subsubsection{Efficiency Analysis.}
\begin{wraptable}{r}{0.5\textwidth}
\centering
\vspace{-3em}
\caption{Inference Latency.}
\fontsize{8pt}{9pt}\selectfont
\setlength{\tabcolsep}{4pt}
\vspace{1em}
\begin{tabular}{l | c}
\toprule
\colorbox{orange!15}{Online}/\colorbox{green!15}{Offline} Method & \textbf{QA Latency} \\
\midrule
\cellcolor{green!15}Qwen2.5-VL-7B         & $0.54s$ \\
\cellcolor{green!15}Qwen2.5-VL-7B w/CoT   & $5.30s$ \\
\cellcolor{green!15}Video-R1 w/CoT        & $8.80s$ \\
\midrule
\cellcolor{orange!15}VideoLLM-online-8B    & $0.38s$ \\
\cellcolor{orange!15}Dispider-7B           & $1.10s$ \\
\cellcolor{orange!15}VST-3B (Ours)        & $0.53s$ \\
\cellcolor{orange!15}VST-7B (Ours)       & $0.56s$ \\
\cellcolor{orange!15}VST-32B (Ours)        & $1.40s$ \\
\bottomrule
\end{tabular}
\vspace{-1em}      
\label{tab:lantency}
\end{wraptable}

We compare the QA latency of several offline and online methods under the same experimental setup. All measurements are conducted on VideoHolmes, as shown in \cref{tab:lantency}. Models without CoT directly output the final answer. Benefiting from our query-ahead streaming think mechanism, VST maintains significantly lower response latency. Moreover, streaming think is executed asynchronously before the query and finishes within the clip inter-arrival interval, so its computation is amortized over playback rather than added after the query. As a result, it does not increase the real-world end-to-end inference time.

In our implementation, streaming thoughts are generated asynchronously every 16--32 seconds and take 7.0 seconds on average, with a P99 latency of 11.2 seconds. Since this is shorter than the minimum trigger interval, the overhead is amortized over playback. If interrupted by a query or a new step, VST uses the latest completed memory state. Therefore, VST adds background computation but not post-query response latency.

\subsubsection{Case Study.}

\begin{figure}[t]
    \centering
    \includegraphics[width=\linewidth]{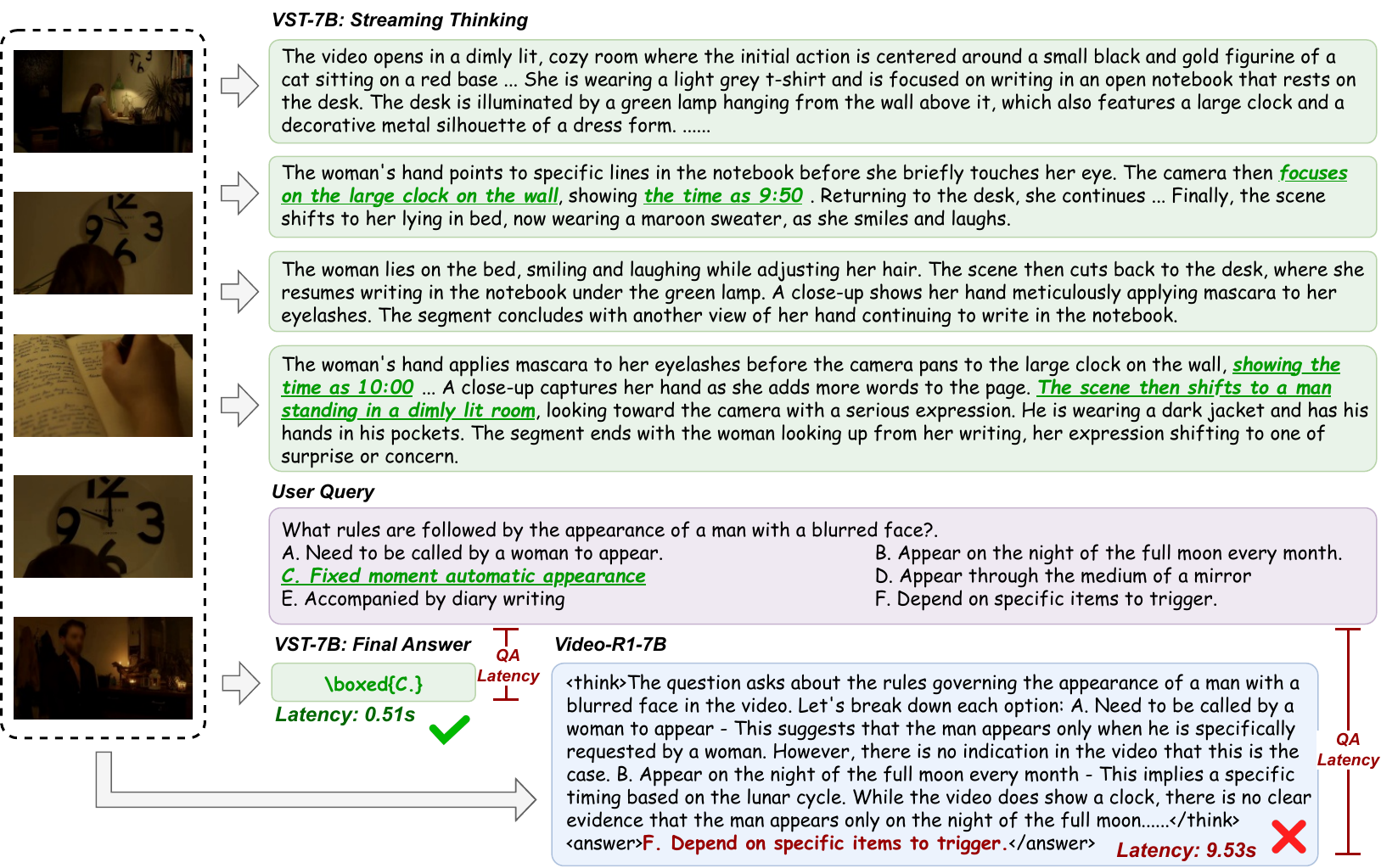} % Placeholder for the uploaded image
    \caption{\textbf{Case Study from VideoHolmes.} We compare VST-7B with Video-R1-7B. VST-7B processes the video stream and performs streaming thinking before the query, then answers directly once the query arrives. In contrast, Video-R1-7B generates CoT after the query, resulting in higher QA latency. VST-7B achieves better performance with lower QA latency in this example.}
    \label{fig:case}
\end{figure}

\cref{fig:case} presents a case study from VideoHolmes. The query requires temporal reasoning over disjoint segments, specifically aligning repeated visualizations of a wall clock with the subsequent appearance of a ``blurred-face man''. The baseline Video-R1-7B, which relies on post-query thinking, fails to capture these dispersed temporal cues due to the difficulty of attending to specific evidence across a long context. Consequently, it hallucinates a spurious correlation involving object interactions, leading to a logical error. Furthermore, this retrospective reasoning incurs a significant latency of 9.53s. In contrast, VST-7B employs streaming thinking to continuously update its evidence (e.g., timestamps and event triggers) as the video memories. This pre-query evidence accumulation allows VST to correctly deduce the time-based rule and, by shifting the reasoning burden to the streaming phase, drastically reduces the response latency to 0.51s. This comparison demonstrates that pre-query streaming thinking simultaneously enhances reasoning robustness and system responsiveness.

We further examine typical failure cases of VST during inference. While streaming thinking improves evidence accumulation, the remaining errors are mainly associated with imperfect memory construction. Specifically, VST may store visually salient but query-irrelevant details, compress early evidence too aggressively, miss fine-grained temporal spans, or fail to associate weak cues across distant events. These observations suggest that more fine-grained and faithful memory construction, especially for preserving subtle temporal evidence and cross-event relations, is a promising direction for future work.

\section{Related Work}
\label{related_work}

\subsubsection{Streaming Video Understanding.}
Streaming video understanding processes continuous visual inputs of indeterminate length. Unlike offline methods, the lack of global sampling and restricted context windows poses significant challenges for VideoLLMs. Some existing methods attempt to retain extended video information within limited context lengths through real-time visual token compression~\cite{song2024moviechat, chen2024videollm, qian2024streaming, yao2025timechat, zengstreamforest,anonymous2026streamingvlm}. 
For instance, rLiVS reduces streaming token redundancy by recurrently pruning attention-uninformative visual tokens, echoing broader attention-reuse strategies for efficient visual processing~\cite{dorovatas2026recurrent,yin2026mstar}.
Others incorporate external memory mechanisms to recall historical information via similarity ranking~\cite{zhang2025flash,qian2025dispider,distreaming,yang2025streammem}. However, these methods rely on static heuristics, lacking autonomous memory management and the ability to perform complex, multi-step reasoning. To bridge this gap, recent studies have introduced streaming and long-context reasoning methods into video understanding~\cite{wen2026eventmemagent,wang2026think}. Along this line, this paper proposes \textit{\textbf{V}ideo \textbf{S}treaming \textbf{T}hinking}~(VST), which introduces an online thinking process that evolves with the video stream. By coupling autonomous memory management with in-depth instruction analysis, VST enables models to transcend short-range perception and achieve robust streaming intelligence.

\subsubsection{VideoLLMs Test-Time Scaling.}
Following the breakthrough of test-time scaling and chain-of-thought in LLMs~\cite{wei2022chain,guo2025deepseek,tong2025streamingthinker}, recent VideoLLMs have adopted supervised fine-tuning (SFT) to mimic expert reasoning trajectories~\cite{hannan2025revisionllm,han2025videoespresso} or utilized R1-style reinforcement learning (RL) to enhance task performance~\cite{feng2025video,yan2025videochatr,wang2025videorft,chenscaling,li2025revisor,jiang2026imagine}. Despite these advances, existing post-training research remains predominantly confined to offline video understanding. The exploration of reasoning within streaming contexts, particularly regarding long-horizon cognitive capabilities, remains a critical yet neglected frontier. In this paper, we introduce a unified SFT and RL framework for streaming video understanding. Our method achieves a synergistic balance between real-time responsiveness and sophisticated reasoning, enabling autonomous memory management and in-depth analysis of evolving video streams.
\section{Conclusion}
In this paper, we propose \textit{\textbf{V}ideo \textbf{S}treaming \textbf{T}hinking} (VST), a new paradigm for streaming video understanding that introduces a synchronized stream of logical inference with real-time responsiveness. VST enables a \textit{thinking-while-watching} mechanism that performs reasoning over incoming clips during streaming. We further develop a post-training recipe (VST-SFT and VST-RL) and an automated data synthesis pipeline based on video knowledge graphs to produce streaming-thought supervision. Empirically, VST delivers strong performance across multiple online and offline video understanding benchmarks, and yields consistent gains across VideoLLMs ranging from 3B to 32B parameters, indicating that the approach generalizes across model scales. Overall, our study establishes VST as a practical test-time scaling approach for streaming scenarios, simultaneously enabling explicit CoT generation and real-time responsiveness.

\subsubsection{Limitation and Future Works.}
While the computation of streaming thoughts can be scheduled in parallel with incoming video clips, the additional LLM token consumption is still non-negligible. A promising direction is to explore \emph{latent reasoning} to enable more token-efficient streaming thinking. Moreover, VST primarily focuses on text-guided memory management, which is orthogonal to existing streaming visual memory mechanisms. Investigating their combination and potential synergy is an interesting avenue for future work.

% ---------------------------------------------------------------
% Acknowledgements
% Camera-ready version should include acknowledgements, if any.
% Use an unnumbered section.
% ---------------------------------------------------------------
\section*{Acknowledgements}
This work was done during the research internship of Yiran Guan and Liang Yin at Xiaomi Inc.

\noindent This work is supported by the NSFC (62225603).

% ---------------------------------------------------------------
% Bibliography
% ---------------------------------------------------------------
\bibliographystyle{splncs04}
\bibliography{reference}

% ---------------------------------------------------------------
% Appendix / Supplementary Material
% ---------------------------------------------------------------
% Be careful: whether appendix can be included in the main camera-ready PDF
% depends on ECCV's final submission instructions.
% If appendix is intended as supplementary material, submit it separately
% instead of including it in the main proceedings PDF.
%
\clearpage
\section{Details of VST Inference}

\subsection{Inference Prompt}
We detail the inference prompts used for\textit{\textbf{V}ideo \textbf{S}treaming \textbf{T}hinking} (VST). Both the training and inference phases strictly follow this format. As discussed in the Method part, the LLM generates two distinct types of responses: 1) intermittent streaming inference as the video progresses, which is conditioned primarily on past memory and the current video clip (\cref{tab:template}, top); and 2) final answer generation upon receiving the user query, which is conditioned on the accumulated memory, the current video clip, and the specific question (\cref{tab:template}, bottom).
\begin{table}[h]
  \centering
  \caption{Template of VST inference for streaming thinking (top part) and final answer generation (bottom). Curly-brace placeholders {\ttfamily\{\}} will be replaced with actual content. \textcolor{violet}{\ttfamily \{Memory\}} denotes the historical outputs from previous streaming thinking, comprising the corresponding timestamps and generated content, while \textcolor{teal}{\ttfamily \{VideoClip\}} represents the incoming video stream.}
  \ttfamily 
  \renewcommand{\arraystretch}{1.1}
  \begin{tabular}{@{}p{\textwidth}@{}}
    \toprule
    % ------------------------ Context-processing prompt ------------------------
    \raggedright
    [System] \newline
    You are a Streaming Video Analyst. \newline
    \textcolor{violet}{\{Memory\}} \newline 
    \{TimeStamp\} \textcolor{teal}{\{VideoClip\}} \tabularnewline 
    \midrule
    % ------------------------ Answer-generation prompt ------------------------
    \raggedright
    [System] \newline
    You are a Streaming Video Analyst. \newline
    \textcolor{violet}{\{Memory\}} \newline
    \{TimeStamp\} \textcolor{teal}{\{VideoClip\}} \newline 
    \{QueryTime\} Based on the provided \textcolor{violet}{Video Memory} and the \textcolor{teal}{Current Video Clip}, answer the following Problem. \newline
    \{Problem\} \newline
    Output the final answer in \verb|\boxed{}| \newline
    \textbf{Your answer:} \tabularnewline
    \bottomrule
  \end{tabular}
  \normalfont 
  \label{tab:template}
\end{table}

\subsection{Streaming Inference Pipeline}
\begin{figure*}[t]
    \centering
    \includegraphics[width=1.0
    \textwidth]{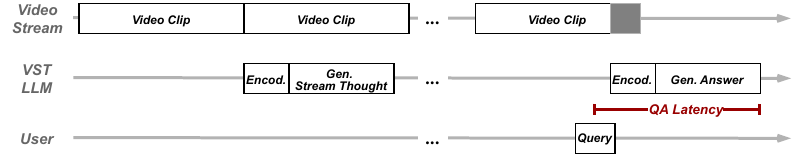}
    \caption{\textbf{The streaming inference pipeline of VST.} By generating stream thoughts for incoming video clips before a user query arrives, VST effectively hides reasoning latency and enables rapid QA responses.}
    \label{fig:stream_inference}
\end{figure*}

\cref{fig:stream_inference} illustrates the streaming reasoning pipeline of VST. Prior to receiving the user query, we conduct a streaming thinking process for each video clip, ensuring the output is generated before the subsequent clip arrives. Consequently, our method effectively utilizes the natural waiting time inherent in real-world video streams. This enables a rapid response once the user poses a question, where the QA latency is defined as the time elapsed from the user's query submission to the LLM's response.

\section{Details of VST-SFT / RL Training}
\paragraph{VST-SFT Hyperparameters.} To train a streaming model capable of continuous thinking and ensure reproducibility, we report the hyperparameters for the VST-SFT phase. We sample up to 384 frames per video and limit the maximum video pixels to 19,267,584, capping video tokens at 24K to reserve 8K context for language and reasoning. The model is trained for 1 epoch with a learning rate of 5e-6 and 8 gradient accumulation steps.

\paragraph{VST-RL Hyperparameters.} For the reinforcement learning phase, we employ the GRPO algorithm to optimize the model. The maximum prompt length is limited to 11,000 tokens, with 1,000 reserved for the maximum generated response length. During the rollout phase, we generate 8 candidate responses per prompt using vLLM with a temperature of 1.0 and a top-p of 0.98. The model is trained for 1 epoch with a global training batch size of 256 and a PPO mini-batch size of 64. We use a learning rate of 5e-7 with 20 warmup steps for the actor model. To ensure stable optimization and manage memory efficiently, we freeze the vision tower, set the KL penalty coefficient to 0.001, and leverage Fully Sharded Data Parallel (FSDP) with both parameter and optimizer offloading.

\begin{table}[t!]
  \centering
  \scriptsize % 进一步缩小字号到 scriptsize，若觉得太小可改回 \footnotesize
  % \footnotesize % 进一步缩小字号到 scriptsize，若觉得太小可改回 
  \ttfamily 
  \renewcommand{\arraystretch}{1.1} % 稍微收紧行间距
  \caption{Templates of prompts used for data generation. The table presents three distinct prompts: Video Knowledge Graph Generation (top), Intermediate Chain-of-Thought (CoT) Generation (middle), and Question-Answer (QA) Generation (bottom). Curly-brace placeholders \textcolor{teal}{\ttfamily \{...\}} will be replaced with actual segment-specific content during generation.}
  \label{tab:data_generation_prompts}
  \begin{tabular}{@{} >{\raggedright\arraybackslash}p{\textwidth} @{}}
    \toprule
    % ------------------------ Knowledge Graph Generation Prompt ------------------------
     You are a \textbf{Visual Scene Analyst} specializing in dense scene graph generation. Your goal is to map \textbf{ALL} physical relationships in the video segment, not just human actions. \\
    {[}CURRENT TIMELINE{]} Segment \textcolor{teal}{\{\#step\_index\}}: \textcolor{teal}{\{start\_time\}}s to \textcolor{teal}{\{end\_time\}}s. \\
    {[}CONTEXTUAL DATA{]} 1. \textbf{Entity Registry}: \textcolor{teal}{\{known\_entities\_str\}} \\
    {[}CRITICAL VISUAL EXTRACTION RULES{]} 1. \textbf{NO "HUB-AND-SPOKE" BIAS (CRITICAL)}: Do not make the human protagonist the subject of every relation. Extract edges where neither the subject nor the object is a person. 2. \textbf{Object-to-Object Relationships (MANDATORY)}: Look for Spatial Relations, Containment, and Passive Interactions. 3. \textbf{Visual Entity Identification}: Identify objects using specific visual descriptors. NO PRONOUNS. 4. \textbf{Action \& State Verbs}: Use active verbs for humans and spatial/state verbs for objects. 5. \textbf{Description}: Describe the scene layout and object states, not just the human's movement. \\
    {[}OUTPUT FORMAT{]} Return ONLY a JSON object: \texttt{\{"events": [\{"subject": "...", "relation": "...", "object": "...", "description": "..."\}]\}} \\
    \midrule
    
    % ------------------------ Intermediate CoT Generation Prompt ------------------------
    \textbf{Current Video Segment}: \textcolor{teal}{\{current\_time\_range\}} \\
    \textbf{Task}: Provide a incremental update of the visual scene/action/details in this segment. \textbf{Focus on Progress (Delta)} of visual elements. \\
    \textbf{Context \& Entities}: Focused Entities: \textcolor{teal}{\{entity\_text\}} (Integrate these naturally if they are actively involved in the current action. No full repetition each time; pronouns prohibited.) \\
    \textbf{Strict Constraints}: 1. \textbf{Focus on Dynamics}: Describe WHAT is happening now with details. If the scene is static, be very concise. 2. \textbf{NO restate}: Do NOT restate the action/scene/details from the previous segment's end. Skip the action stated previously or state the changes. NEVER repeat. 3. \textbf{Minimalist Reference}: For entities already present/mentioned, use pronouns or minimal descriptors. 4. \textbf{Language Variety}: Avoid repetitive sentence structures. 5. \textbf{No Redundancy}: Do not repeat the timestamp or information from the previous context. \\
    \textbf{Output}: \\
    \midrule

    % ------------------------ QA Generation Prompt ------------------------
    You are a \textbf{Cognitive Video Intelligence Engine}. Your task is to synthesize a \textbf{Practical Deep Reasoning Question-Answer Pair} based on the provided visual information from video segments and the \textbf{Event Reasoning Path}. \\
    \textbf{=== CORE DEFINITION (MUST FOLLOW FIRST) ===} 1. \textbf{Event Reasoning Path Nature}: The provided video segments are ordered by logical reasoning relevance, NOT by their original chronological time sequence. 2. \textbf{Time Reference Rule}: All time mentions must be based on the explicit time interval of each node. 3. \textbf{Multi-hop Foundation}: The reasoning must rely on logical connections between segments. \\
    \textbf{=== MANDATORY INSTRUCTIONS ===} 1. \textbf{Strict Multi-hop Reasoning}: The question MUST require integrating information from multiple video segments in the reasoning path. 2. \textbf{Natural Language Constraint (CRITICAL)}: DO NOT use the words "Step", "Clip index", "Path node". Refer to segments using their time intervals or event descriptions. 3. \textbf{Reasoning Dimensions}: Construct the reasoning chain using one or more of the following logic types: \textcolor{teal}{\{dimensions\_str\}} 4. \textbf{Practicality Requirement}: The question must be meaningful for understanding the visual content of the video's narrative, intent, or physical logic. \\
    \textbf{=== OUTPUT FORMAT (JSON ONLY, NO EXTRA TEXT) ===} \texttt{\{"question": "...", "cot": "...", "answer": "...", "reasoning\_type": "..."\}} \\
    \bottomrule
  \end{tabular}
\end{table}

\section{Details of VST Data Generation}
Table \ref{tab:data_generation_prompts} presents the prompt templates utilized for our data generation pipeline. The entire data synthesis process strictly adheres to these formats to ensure high-quality, consistent annotations. As discussed in the Data Generation section, the LLM generates three distinct types of outputs: 1) video knowledge graph construction, which is conditioned on the current video segment and known entities to map dense physical relationships and object states (Tab. \ref{tab:data_generation_prompts}, top); 2) intermediate chain-of-thought (CoT) generation, which is conditioned on the specific time range and focused entities to provide incremental updates of the visual progress (Tab. \ref{tab:data_generation_prompts}, middle); and 3) multi-hop QA generation, which is conditioned on the constructed event reasoning path and visual evidence to synthesize practical reasoning question-answer pairs (Tab. \ref{tab:data_generation_prompts}, bottom).

\end{document}